\documentclass{article}

\usepackage{arxiv}

\usepackage{fancyhdr}
\date{}  
\makeatletter
\renewcommand{\@date}{}  
\makeatother

\usepackage[utf8]{inputenc}
\usepackage[T1]{fontenc}
\usepackage{hyperref}
\usepackage{url}
\usepackage{booktabs}
\usepackage{amsfonts}
\usepackage{nicefrac}
\usepackage{microtype}
\usepackage{lipsum}
\usepackage{graphicx}
\usepackage{amsmath}
\usepackage{float}
\graphicspath{{./images/}}

\title{IDRIFTNET: Physics-Driven Spatiotemporal Deep Learning for Iceberg Drift Forecasting}

\author{
 Rohan Putatunda \\
  Department of Information Systems\\
  University of Maryland Baltimore County\\
  USA \\
  \texttt{rohanp1@umbc.edu} \\
  \And
 Sanjay Purushotham \\
  Department of Information Systems\\
  University of Maryland Baltimore County\\
  USA \\
  \texttt{psanjay@umbc.edu} \\
  \And
 Ratnaksha Lele \\
  iHARP: NSF HDR Institute\\
  University of Maryland Baltimore County\\
  USA \\
  \texttt{rlele@umbc.edu} \\
  \And
 Vandana P. Janeja \\
  Department of Information Systems\\
  University of Maryland Baltimore County\\
  USA \\
  \texttt{vjaneja@umbc.edu} \\
}

\begin{document}
\maketitle
\begin{abstract}
Drifting icebergs in the polar oceans play a key role in the Earth's climate system, impacting freshwater fluxes into the ocean and regional ecosystems while also posing a challenge to polar navigation. However, accurately forecasting iceberg trajectories remains a formidable challenge, primarily due to the scarcity of spatiotemporal data and the complex, nonlinear nature of iceberg motion, which is also impacted by environmental variables. The iceberg motion is influenced by multiple dynamic environmental factors, creating a highly variable system that makes trajectory identification complex. These limitations hinder the ability of deep learning models to effectively capture the underlying dynamics and provide reliable predictive outcomes. To address these challenges, we propose a hybrid IDRIFTNET model, a physics-driven deep learning model that combines an analytical formulation of iceberg drift physics, with an augmented residual learning model. The model learns the pattern of mismatch between the analytical solution and ground-truth observations, which is combined with a rotate-augmented spectral neural network that captures both global and local patterns from the data to forecast future iceberg drift positions. We compare IDRIFTNET model performance with state-of-the-art models on two Antarctic icebergs: A23A and B22A. Our findings demonstrate that IDRIFTNET outperforms other models by achieving a lower Final Displacement Error (FDE) and Average Displacement Error (ADE) across a variety of time points. These results highlight IDRIFTNET's effectiveness in capturing the complex, nonlinear drift of icebergs for forecasting iceberg trajectories under limited data and dynamic environmental conditions.
\end{abstract}


\section{Introduction}
\begin{figure}[t]
  \centering
  \includegraphics[width=0.50\textwidth]{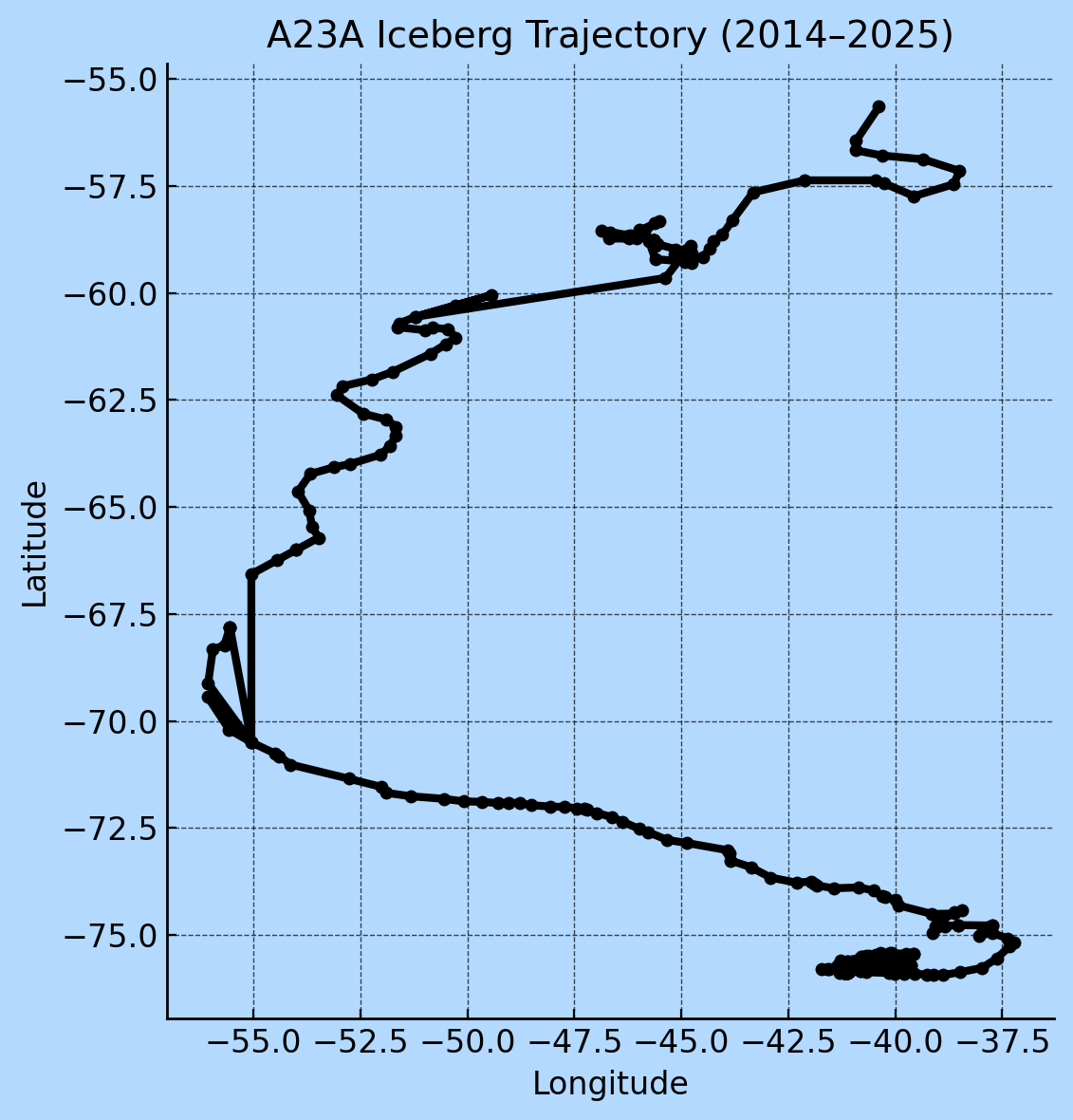}
  \caption{A23A iceberg trajectory from 2014 to 2025 exhibits uncertain drift patterns including zig-zag motion, looping, and stagnation zones.}
  \label{fig:A23A Uncertain Movement Pattern}
\end{figure}

Forecasting the drift trajectories of icebergs has become increasingly important because of its consequences for marine safety, regional climate modeling, and estimation of the freshwater intake into the ocean. Enormous icebergs, especially ones such as those calved from the Antarctic ice shelves can remain afloat and mobile for years, with consequences for regional ocean circulation by  effecting air-sea and freshwater fluxes and local ecosystem dynamics. Predicting iceberg trajectories accurately using a statistical modeling is a challenging task due to the small amount of spatiotemporal data available to develop sophisticated tracking frameworks. Moreover the stochastic and non-linear patterns of drift make it even more challenging to predict trajectories of drifting icebergs ( as shown in Figure \ref{fig:A23A Uncertain Movement Pattern}). Iceberg drift is influenced by several dynamic environmental factors, such as wind and ocean currents, as well as intrinsic properties like the iceberg’s surface area, which modulate its response to these forces. Conventional drift forecasting models, those used by the Canadian Ice Service and the International Ice Patrol have mostly depended on deterministic formulations or empirical heuristics based on averaged environmental influences. One such theory, often referred to as the ``2 wind rule'', suggests that the iceberg drift velocity corresponds roughly to 2$\%$ of the surface wind velocity relative to ocean currents \cite{garrett1985tidal,smith1987dynamic,smith1993hindcasting,bigg1997modelling}. This theory often holds in Arctic settings for smaller bergs, however, this assumption breaks down in the Antarctic, where major tabular icebergs show movement pattern based on both the wind and the ocean current as one of the major impacting factors \cite{wagner2017analytical}. Moreover, models depending just on physics-based formulations often suffer from parameter uncertainty, especially in the estimation of drag coefficients and ocean current profiles \cite{andersson2016moving}.

New directions for iceberg drift forecasting have been opened by recent developments in data-driven modeling. Particularly, hybrid learning methods combining physical knowledge with machine learning have shown success in capturing the dynamics of iceberg motion in the presence of noisy and insufficient data. Yulmetov et al. (2011) \cite{yulmetov2021iceberg} proposed a gradient-boosted tree-based system, explicitly computing coriolis acceleration while learning hydrodynamic drag accelerations from GPS-tagged iceberg trajectories. Particularly in environments with limited in situ measurements, this semi-empirical method compares well to pure ML or physics-based methods for short forecast lead times (up to 24 hours). Similarly, Wesche et al. \cite{wesche2014ice} used Synthetic Aperture Radar (SAR) and edge-detection techniques to track iceberg placements and trajectories across the Weddell Sea, therefore verifying the relevance of current-dominated drift areas. However, these approaches often struggle to scale beyond short-term forecasts due to limited data, which ranges from days to weeks and exhibits complex environmental dependencies. This highlights the need for robust hybrid models that integrate physical knowledge with learning from limited spatiotemporal observations. Additionally, these approaches are also limited to short temporal windows (24 hours) and usually depend on constant access to GPS-tagged observations, which are impractical for larger non-geotagged icebergs. To address these challenges, we present IDRIFTNET, a hybrid deep learning architecture combining data-driven learning with physical modeling, to tackle these challenges. 

\noindent
The core contributions of our work are structured in three folds to develop the IDRIFTNET algorithm to predict the future iceberg drift position under limited spatio temporal data scenario with the nonlinear movement pattern of these icebergs:

\begin{itemize}
    \item[(a)] A residual physics module based on Wagner et al. 2017's \cite{wagner2017analytical} analytical drift formulation that guides learning with domain-specific knowledge;
    \item[(b)] A Rotate Block that enlarges the temporal receptive field via channel-wise time shifts, capturing delayed environmental effects;
    \item[(c)] A Gabor-Spectral network that models temporal dependencies by transforming inputs into the frequency domain using FFT to capture global patterns, and applying Gabor filters to isolate localized frequency bands, thereby enabling the learning of both smooth long-range trends and abrupt short-term variations in iceberg drift.
\end{itemize}

\noindent

The remainder of this paper is structured as follows: Section 2 presents the related work associated with this study; Section 3 describes the overall methodology and the proposed architecture; Section 4 discusses the experimental setup, results, quantitative evaluations, ablation studies including the forecasting visualization; Section 5 reports the conclusion and future work.

\section{Related Work}
We describe related work in two core areas: (1)  physics based approaches in modeling the iceberg drift prediction, and (2) state of the art deep learning models that are widely used in various applications related to trajectory prediction.

\subsection{Physics-Based Drift Models}
To forecast the iceberg movements, a classical physics-based model usually takes force balance equations, including wind drag, ocean current drag, coriolis force, and pressure gradients into consideration as highlighted in the early contributions for modeling iceberg motion with the momentum-based formulations as proposed in the work by Bigg et al. \cite{bigg1997modelling}, Gladstone et al. \cite{gladstone2001iceberg}, and Martin et al. \cite{martin2010parameterizing}. These models often limit their accuracy under some oceanic situations by assuming quasi-steady state dynamics for the iceberg, neglecting acceleration, vertical shear effects, and wave radiation forces. These works are related to the two percent rule and are often applicable to small icebergs \cite{garrett1985tidal}. To overcome the challenge associated with iceberg drift estimation by these methods, Wesche et al. \cite{wesche2014ice} applied a wind-driven analytical model (S-IB) in their iceberg monitoring framework using SAR imagery. Their model is based on formulations by Crépon et al. \cite{crepon1988drift} and includes basic force components such as ocean and wind drag, while neglecting sea-ice interactions and vertical shear. Although this approach was effective for short-term drift estimation in the Weddell Sea, it demonstrated limited accuracy in regions with complex sea-ice dynamics or strong environmental variability. The drift paths modeled by S-IB often deviated from observed iceberg trajectories when wind forcing was underestimated or when sea-ice interactions played a significant role. To address the overall challenges,   Wagner et al. \cite{wagner2017analytical} proposed a strong analytical model derived from basic principles of Newtonian law that integrates iceberg area, wind and water drag as well as coriolis effects in order to solve these constraints. This approach additionally includes a scale-aware dimensionless parameter to identify whether drift is wind or current-dominated, therefore offering great physical insight with minimum empirical adjustment that is related to parameter tuning. Its physical interpretability and analytical tractability makes it as a perfect fit for hybrid learning architectures.

\subsection{Deep Learning Models for Trajectory Forecasting}
Recent advancements in trajectory prediction have introduced a variety of neural architectures to better capture the spatiotemporal complexity and multimodal nature of agent dynamics. Trajectron++, proposed by Salzmann et al. \cite{salzmann2020trajectron++}, is a graph-structured recurrent model that incorporates dynamic feasibility and semantic maps, enabling predictions conditioned on both agent behavior and environmental context. In another notable work, Mangalam et al. proposed the Predicted Endpoint Conditioned Network (PECNet) \cite{mangalam2020not}, which formulates trajectory forecasting as an endpoint-conditioned process—first predicting plausible future destinations and then generating socially compliant paths conditioned on those endpoints.

Neural Motion Message Passing (NMMP), introduced by Hu et al. \cite{hu2020collaborative}, applies neural message passing over actor interaction graphs, learning directed motion features between entities for more interpretable and cooperative trajectory forecasting. AgentFormer, introduced by Yuan et al. \cite{yuan2021agentformer}, leverages a transformer-based architecture with agent-aware attention, jointly modeling temporal evolution and social interactions by preserving both time and agent identity in a unified sequence representation.

To improve interpretability, Xu et al. proposed the Memory-based Intention Prediction Network (MemoNet) \cite{xu2022remember}, which integrates a retrospective memory mechanism that retrieves instance-based knowledge from stored past-future pairs, emulating cognitive memory systems to enhance destination prediction.The follow-up work, Group-Aware Trajectory Prediction Network (GroupNet), also proposed by Xu et al. \cite{xu2022groupnet}, extends this concept by employing a multiscale hypergraph structure to capture both pairwise and group-wise relations. It reasons over interaction categories and strengths, enabling more expressive relational modeling and yielding consistent improvements across multiple prediction frameworks.

Motion Indeterminacy Diffusion (MID), proposed by Gu et al. \cite{gu2022stochastic}, models future uncertainty through a reverse diffusion process, where a parameterized Markov chain gradually refines noisy trajectories into determinate predictions. Non-Probability Sampling Network (NPSN), introduced by Bae et al. \cite{bae2022non}, improves sampling efficiency in stochastic models using Quasi-Monte Carlo strategies and a learnable sampling module, thereby generating diverse yet realistic trajectories with reduced variance.In the domain of stochastic modeling, Leapfrog Diffusion, as proposed by Mao et al. \cite{mao2023leapfrog}, accelerates traditional diffusion-based prediction by introducing a trainable initializer that bypasses redundant denoising steps, allowing for real-time inference while retaining trajectory diversity.

Despite the increasing popularity of deep learning-based trajectory models, two main challenges remain unresolved. First, most of these models are trained on data from applications where the degree of nonlinearity is comparatively lower than that found in iceberg drift data. Second, the convergence of complex neural networks is frequently hindered by the limited amount of data availability.

To address these challenges, we present IDRIFTNET, a hybrid deep learning model that integrates residual physics-informed corrections within a rotated spectral-temporal forecasting framework for robust predictions.

\section{Methodology} \label{sec:methodology}
\begin{figure*}[t]
  \centering
  \includegraphics[width=1\textwidth]{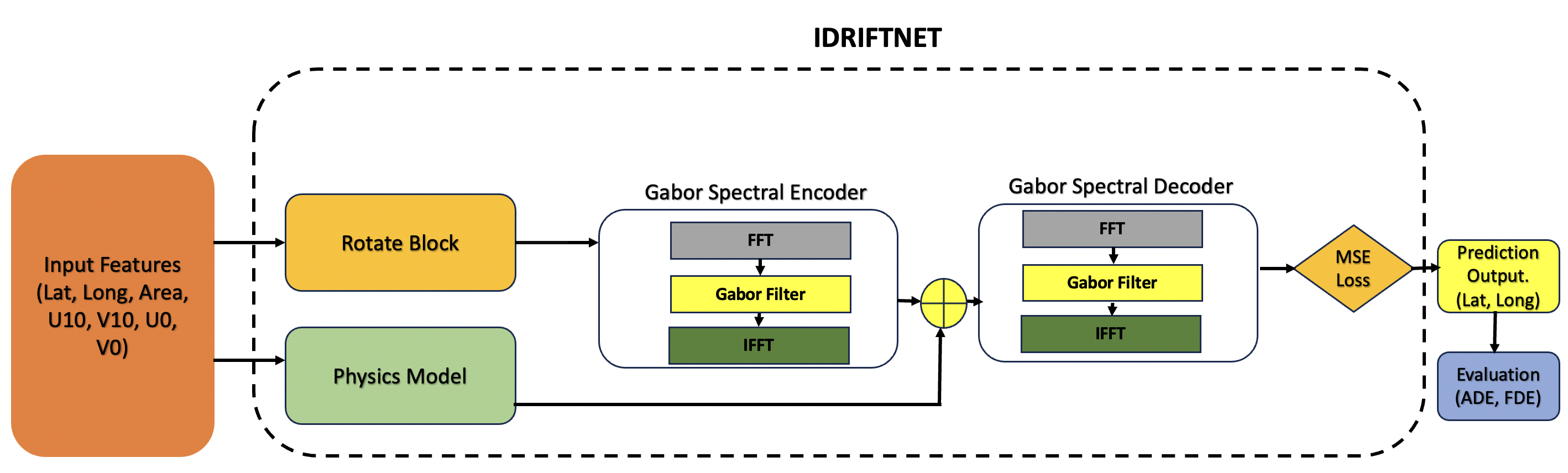} 
  \caption{Overview of the proposed IDRIFTNET Architecture}
  \label{fig:gaborGSN_architecture}
\end{figure*}

To model the spatiotemporal dynamics of iceberg drift with physical knowledge and data-driven generalization, we propose a hybrid neural architecture called IDRIFTNET that is shown in Figure 2. The model combines three core components: (i) a residual physics-informed drift formulation, (ii) rotate block for enhanced receptive field, (iii) Gabor-based spectral neural network, these three core components make the IDRIFTNET a robust model for iceberg drift prediction. We describe each component in detail.

\subsection{Physics-Based Drift Modeling and Residual Learning}
To incorporate domain knowledge of geophysical forces driving iceberg motion, we begin with an analytical formulation derived from Wagner et al. 2017's physics-based drift model \cite{wagner2017analytical}. This formulation models the net drift of an iceberg as a function of wind, ocean current, Coriolis force, and iceberg geometry.

Let $\mathbf{e}_t = [u_{10}, v_{10}, u_o, v_o, A]$ be the vector of environmental variables at time $t$, and $\mathbf{x}_{t-1} = [\phi_{t-1}, \lambda_{t-1}]$ denote the iceberg’s latitude and longitude at the previous timestep. The physical model $\mathcal{F}_{\text{phys}}$ proceeds through the following steps:

\noindent{ \it Wind and Ocean Vector} are defined as, wind vector $\mathbf{v}_a = [u_{10}, v_{10}]$ and ocean current vector $\mathbf{v}_w = [u_o, v_o]$.

\noindent{ \it Iceberg Length Scale:} Assuming the iceberg has area $A$ in square meters, the characteristic length is estimated as:
\begin{equation}
    S = \frac{A}{2 \sqrt{A}}, \quad \text{where } A = A_{\text{km}^2} \times 10^6
\end{equation}

\noindent{ \it Non-Dimensional Drift Parameter:} A scale parameter $L$ is introduced to characterize the relative influence of wind to Coriolis force as follows:
\begin{equation}
    L = \frac{g C_w \|\mathbf{v}_a\|}{\pi f S}, \quad \text{where } f \text{ is the Coriolis parameter}
\end{equation}
with $g$ the gravity constant, $C_w$ the water drag coefficient, and $f$ varying with latitude.

\noindent{ \it Wind Response Coefficients:} is defined based on $L$, compute drift coefficients:
\begin{align}
    a &= \frac{1}{2L^3}(\sqrt{1 + 4L^4} - 1), \\
    b &= \frac{1}{\sqrt{2} L^3} \sqrt{(1 + L^4)\sqrt{1 + 4L^4} - 3L^4 - 1}
\end{align}

\noindent{ \it Wind-Induced Drift Term:} is the effective wind-induced motion and is given by:
\begin{equation}
    \mathbf{v}_{\text{wind}} = -a \cdot \mathbf{v}_a^\perp + b \cdot \mathbf{v}_a
\end{equation}
where $\mathbf{v}_a^\perp$ is the 90-degree rotation of $\mathbf{v}_a$.

\noindent{ \it Net Iceberg Velocity:} The total drift velocity of an iceberg is expressed as:
\begin{equation}
    \mathbf{v}_i = \mathbf{v}_w + \gamma \cdot \mathbf{v}_{\text{wind}}
\end{equation}
where $\mathbf{v}_w$ denotes the ocean current velocity and $\mathbf{v}_{\text{wind}}$ is the wind velocity vector at 10 meters above sea level. The parameter $\gamma$ is a dimensionless empirical coefficient that encapsulates the relative contribution of wind forcing to iceberg drift, accounting for air drag, iceberg geometry, and wind-induced surface stress. This formulation follows standard geophysical modeling conventions used in large-scale iceberg drift simulations.

\noindent{ \it Geolocation Update:} The forecasted coordinates are updated as:
\begin{equation}
    \hat{\phi}_t^{\text{phys}} = \phi_{t-1} + \delta \cdot v_{i,\text{lat}}, \quad
    \hat{\lambda}_t^{\text{phys}} = \lambda_{t-1} + \delta \cdot v_{i,\text{lon}}
\end{equation}
where $\delta$ is a unit conversion factor (e.g., converting m/s to degrees per timestep).

\noindent
This physics-based function $\mathcal{F}_{\text{phys}}(\mathbf{e}_t, \mathbf{x}_{t+1})$ is implemented in TensorFlow and remains fully differentiable, allowing end-to-end integration with neural networks.

\subsubsection*{Residual Learning Connection to the Deep Network:} While the analytical model captures the dominant geophysical forces, it is limited in several ways. Ocean current fields and drag coefficients may be imprecisely known or measured. Fine-scale nonlinear interactions (e.g., iceberg rotation, sea-ice drag) are not explicitly modeled.

To mitigate these shortcomings, we adopt a residual learning approach that augments the physical prediction with a data-driven correction. The final trajectory prediction is defined as:
\begin{equation}
    \hat{\mathbf{y}}_t = \underbrace{\mathcal{F}_{\text{phys}}(\mathbf{e}_t, \mathbf{x}_{t-1})}_{\text{Physics-Based Prediction}} + \underbrace{\mathcal{F}_{\text{net}}(\mathbf{X}_{t-L:t}, \mathbf{e}_t)}_{\text{Neural Residual Correction}}
\end{equation}
where $\mathcal{F}_{\text{net}}$ is a Gabor-based spectral network that receives the historical trajectory sequence $\mathbf{X}_{t-L:t}$ and current environmental input $\mathbf{e}_t$.

The model learns to approximate the residual error $\mathbf{r}_t = \mathbf{y}_t - \hat{\mathbf{p}}_t^{\text{phys}}$ between the physical prediction and ground truth:
\begin{equation}
    \mathbf{r}_t = \mathbf{y}_t - \hat{\mathbf{p}}_t^{\text{phys}}
\end{equation}

This residual learning strategy, inspired by Sun et al. ~\cite{sun2019combining}, enables the deep learning network to focus solely on the discrepancy between physics and ground truth. This not only reduces the model's search space and improves generalization in data-scarce regimes, but also promotes interpretability by clearly separating learned behavior from physically grounded dynamics.

\subsection{Temporal Encoding via Rotate Block}

To enhance the model’s sensitivity to temporal dependencies in environmental sequences, we integrate a Rotate Block ~\cite{khalitov2022chordmixer}. This module performs channel-wise cyclic shifts on each feature dimension along the temporal axis using a dyadic pattern:

\begin{equation}
\mathbf{X}_{\text{rot}}[:, :, i] = \text{roll}(\mathbf{X}[:, :, i], \Delta_i), \quad \Delta_i = 2^{\lfloor i \cdot \log_2 L / d \rfloor}
\end{equation}

Here, $L$ is the input sequence length, $d$ is the number of feature channels, and $\Delta_i$ denotes the shift amount for the $i$-th feature channel. This operation distributes temporal offsets non-uniformly across channels, introducing inductive bias for capturing lagged dependencies at varying temporal scales.

In the context of modeling iceberg drift, such temporal offsets are particularly valuable. Environmental forces such as wind, ocean currents, and iceberg area exhibit delayed or asynchronous effects on motion. For instance, a sudden change in wind may affect the drift trajectory several time steps later, depending on iceberg inertia and oceanic damping. The Rotate Block helps the network to capture such delayed causal signals by dilating the receptive field across time without increasing model complexity or adding parameters. By enabling learned feature channels to attend to temporally shifted representations, the Rotate Block strengthens IDRIFTNET’s ability to model non-local temporal dynamics. This ultimately improves the fidelity of long-range trajectory forecasting and helps to encode temporal dependencies or abrupt changes in forcing variables that influence iceberg motion.

\subsection{Learning with Gabor Spectral Network (GSN)}

One of the core architectural innovations of the IDRIFTNET model is the integration of a Gabor Spectral Network (GSN) within an encoder-decoder framework for spatiotemporal sequence learning. Unlike standard temporal models operating solely in the time domain, GSNs transform input sequences into the frequency domain, leveraging Gabor filters to learn global patterns while preserving local temporal structure. This approach is particularly effective in capturing both gradual drifts and sudden shifts in iceberg movement trajectories, which are otherwise difficult to model using conventional convolutional or recurrent architectures.

\paragraph{Input Representation and Preprocessing.} 
We consider a multivariate time-series input $\mathbf{X} \in \mathbb{R}^{T \times d}$, where $T$ is the sequence length and $d=7$ includes geospatial and environmental features: latitude, longitude, ice area, surface wind components ($u_{10}, v_{10}$), and ocean current components ($u_o, v_o$). The input is normalized using a MinMax scaling technique to map values to the $[0, 1]$ range, ensuring stability during training and preserving relative feature magnitudes.

\paragraph{Spectral Encoder Using Gabor Convolution.}
The encoder first transforms the time-domain signal into its frequency-domain representation using a Real FFT (Fast Fourier Transform). For each input-output channel pair $(i,j)$, a trainable Gabor filter is applied in the spectral domain. This filter is defined as:

\begin{equation}
\mathcal{G}_{ijm} = \exp\left(-\frac{(f_m - \mu_{ij})^2}{2\sigma_{ij}^2}\right),
\end{equation}

where $f_m$ represents the $m$-th normalized frequency mode, and $\mu_{ij}, \sigma_{ij}$ are learnable parameters controlling the center frequency and bandwidth. The filtered signal is obtained by multiplying the transformed input $\mathcal{F}(\mathbf{X}_i)$ with the Gabor filter $\mathcal{G}_{ij}$ and a learnable spectral weight matrix $W_{ij}$. The result is transformed back to the time domain using the inverse FFT:

\begin{equation}
\hat{\mathbf{X}}_j = \mathcal{F}^{-1} \left( \sum_{i=1}^{d} \mathcal{F}(\mathbf{X}_i) \cdot \mathcal{G}_{ij} \cdot W_{ij} \right).
\end{equation}

A RotateBlock \cite{khalitov2022chordmixer} is applied to each channel before the spectral convolution, cyclically shifting each feature channel by a dyadic offset, $\Delta_i = 2^{\lfloor i \cdot \log_2 T / d \rfloor}$, to promote multi-scale temporal interactions.

\paragraph{Physics-Informed Residual Decoder.}
The decoder predicts residual corrections over a physics-based forecast. Given a contextual feature vector composed of environmental dynamics and encoded features, $\mathbf{z} \oplus \mathbf{tw}$, where $\mathbf{z}$ is the encoder output and $\mathbf{tw} \in \mathbb{R}^7$ contains wind, ocean current, area values, and previous location (latitude and longitude), we define the residual as:

\begin{equation}
\Delta \mathbf{y} = \text{Decoder}_{\text{GSN}}([\mathbf{z}, \mathbf{tw}]).
\end{equation}

The decoder applies the spectral filtering mechanism for correction. The final output is produced by combining the physics-based drift model $\mathbf{y}_{\text{phys}}$ with the learned correction:

\begin{equation}
\mathbf{y}_{\text{final}} = \mathbf{y}_{\text{phys}} + \Delta \mathbf{y}.
\end{equation}

We integrate a drift model based on Wagner et al. 2017's physics-based formulation into our deep learning network. This physics model serves as a baseline trajectory forecast, to which the decoder adds a data-driven correction, enabling hybrid forecasting through which the model gains the physical knowledge and learned temporal dynamics for forecasting future iceberg drift.

Gabor filters offer localized frequency representations, maintaining time-frequency coherence—an advantage over standard Fourier-based convolution, which lacks temporal resolution. This compact and efficient formulation captures long-range dependencies without requiring deep stacks of convolution layers or recurrent mechanisms. The encoder compresses historical patterns into expressive spectral features, while the decoder leverages environmental context to refine the iceberg drift predictions. The spectral formulation ensures stability and generalization across varying temporal scales and conditions.
 By integrating rotated Gabor Spectral Networks with physics-informed residual correction, IDRIFTNET offers a principled and efficient solution for forecasting complex iceberg drift trajectories.

\section{Experimental Results}
In this section, we will discuss the data and the preprocessing steps, the implementation details, the evaluation metrics, and the results obtained for this study.

\subsection{Data and Preprocessing}
This study focuses on forecasting iceberg drift trajectories by integrating geophysical position data with environmental forcing variables. Our dataset spans from 2014 to February 21, 2025, and includes records for icebergs A23A and B22A. Each data sample contains a time stamped entry that includes the iceberg's geolocation (latitude and longitude), surface area, and four environmental features: near-surface wind velocity components (u10, v10), and ocean surface current components (uo, vo). These variables were selected based on their direct influence on iceberg motion and melting dynamics \cite{wagner2017analytical}.

The spatial and temporal location data of the icebergs, including their centroid coordinates and surface area, were obtained from the U.S. National Ice Center (USNIC), which routinely tracks Antarctic icebergs using satellite-based observations and publishes the verified geolocations. These observations were stored in csv data format and loaded using pandas. Column names were standardized, and the Date field was parsed to datetime format with normalization to ensure uniformity. The resulting dataframe was sorted chronologically to maintain sequential integrity.

The environmental variables were collected from multiple data repositories. Wind velocity components are collected from the ERA5 climate reanalysis dataset available on the Copernicus Climate Data Store (CDS). These datasets were provided as daily gridded CSV files and were aggregated into daily averages by grouping on the Date field. While ocean current velocity components are obtained from the  Copernicus Marine Environment Monitoring Service (CMEMS), which supplies global ocean physical state variables.

Each file was cleaned by parsing and normalizing the time columns, followed by daily aggregation. After preparing all components, the environmental variables were merged with the USNIC iceberg metadata using an outer join on the Date field. The complete dataset was used for subsequent model training.

\subsection{Implementation}
We developed and trained our model on the Google Colab platform using an NVIDIA Tesla T4 GPU with 16~GB of VRAM. The system was configured with NVIDIA driver version 550.54.15 and CUDA version 12.4 to ensure compatibility with TensorFlow. All experiments were conducted using TensorFlow 2.x, and data preprocessing was performed using widely adopted Python libraries such as Pandas, NumPy, and Scikit-learn. Our proposed model IDRIFTNET, is a physics-integrated deep learning architecture designed to predict iceberg trajectories. It combines analytical physics drift modeling with frequency-aware neural encoding. The model processes input in the form of sliding windows consisting of five consecutive timesteps. Each timestep includes seven normalized features: latitude, longitude, area, surface wind components (u10 and v10), and ocean current velocities (uo and vo). The goal is to predict the iceberg’s latitude and longitude at the next timestep.

Our model IDRIFTNET consists of a Gabor Spectral Network-based encoder-decoder architecture, as detailed in Section \ref{sec:methodology}, with a physics-informed residual branch based on Wagner et al. (2017). The spectral blocks operate in the frequency domain, enabling the network to learn both global and localized temporal patterns, while the rotate layer enhances temporal diversity through cyclic feature shifts. A physics module provides an initial drift estimate, which the decoder refines via learned residual corrections using environmental inputs. All input features are normalized to the [0, 1] range using a MinMax scaler. The model is trained to minimize Mean Squared Error (MSE) between predicted and ground-truth coordinates. We use the Adam optimizer with a learning rate of $1\times10^{-4}$ and apply dropout at a rate of 0.2 after each spectral block to mitigate overfitting. Training proceeds for up to 500 epochs using full-batch updates, with early stopping triggered if the validation loss does not improve for 100 consecutive epochs. The best-performing model, based on validation loss, is checkpointed and used for final evaluation.

\subsection{Autoregressive Inference Strategy}

To evaluate the model’s performance under realistic deployment scenarios, we adopted an autoregressive inference strategy during test time. Starting with an initial input sequence of five consecutive timesteps, the model recursively predicted the next location one step at a time. At each step, the predicted latitude and longitude were combined with the corresponding environmental features to construct the input for the next prediction. For a given $\mathbf{X}_{t:t+4} \in \mathbb{R}^{5 \times d}$ that denote the initial input sequence of normalized features, where $d$ is the number of input variables. The model generates a prediction $\hat{\mathbf{y}}_{t+5} = f_\theta(\mathbf{X}_{t:t+4}, \mathbf{z}_{t+5})$, where $\mathbf{z}_{t+5}$ consists of the environmental drivers like wind, ocean current, and area at time $t+5$ along with the most recent known coordinates from time $t+4$. The new input sequence is then updated by removing the oldest entry $\mathbf{X}_t$ and appending the newly formed feature vector $[\hat{\mathbf{y}}_{t+5}, \mathbf{z}^{(env)}_{t+5}]$ to the sequence, forming $\mathbf{X}_{t+1:t+5}$. This process is repeated iteratively across the entire prediction horizon. This autoregressive rollout mimics real-world sequential forecasting, where future inputs must be inferred from past predictions and known covariates. It also exposes the model to the compounding effect of prediction errors, making the evaluation more robust and realistic.

\subsection{Evaluation Metrics}

To quantitatively assess the accuracy of predicted iceberg trajectories, we adopt two standard metrics widely used in spatiotemporal sequence prediction: Average Displacement Error (ADE) and Final Displacement Error (FDE). These metrics are calculated in both geographic degrees and error distance in meters (converting meters into kilometers) using the geodesic distance based on haversine provided by the geopy library.

\paragraph{Average Displacement Error (ADE):} 
ADE measures the average L2 distance between the predicted trajectory and the ground truth for all prediction time steps. Given a predicted trajectory $\hat{\mathbf{x}}_t = (\hat{\phi}_t, \hat{\lambda}_t)$ and a ground truth trajectory $\mathbf{x}_t = (\phi_t, \lambda_t)$ for $T$ steps, ADE is defined as:

\begin{equation}
\mathrm{ADE} = \frac{1}{T} \sum_{t=1}^{T} \left\| \hat{\mathbf{x}}_t - \mathbf{x}_t \right\|_2
= \frac{1}{T} \sum_{t=1}^{T} \sqrt{(\hat{\phi}_t - \phi_t)^2 + (\hat{\lambda}_t - \lambda_t)^2}
\end{equation}

Here, $\hat{\phi}_t$ and $\hat{\lambda}_t$ denote the predicted latitude and longitude at timestep $t$, while $\phi_t$ and $\lambda_t$ represent the corresponding ground truth values. When evaluated in meters(which we converted into kilometers), this formula is computed using geodesic distances on the Earth's surface rather than the Euclidean distance in degrees.

\paragraph{Final Displacement Error (FDE):}
FDE measures the endpoint accuracy of the predicted sequence. It calculates the L2 distance between the predicted final location and the ground truth final location. Formally, it is given by:

\begin{equation}
\mathrm{FDE} = \left\| \hat{\mathbf{x}}_T - \mathbf{x}_T \right\|_2
= \sqrt{(\hat{\phi}_T - \phi_T)^2 + (\hat{\lambda}_T - \lambda_T)^2}
\end{equation}

This metric is crucial in applications where the final predicted position of the iceberg, which will showcase the importance for downstream decision making, such as iceberg tracking and collision avoidance as a part of maritime safety.

\paragraph{Geodesic Distance Conversion.}
While the above expressions use angular units (degrees), we convert them to real-world units (meters) using the geodesic distance function, which accounts for Earth's curvature. For each timestep $t$, the geodesic distance $d_t$ between the ground truth and predicted point is computed as:

\begin{equation}
\label{eq: geodesic_distance}
d_t = \mathrm{Geodesic} \left( (\phi_t, \lambda_t), (\hat{\phi}_t, \hat{\lambda}_t) \right)
\end{equation}

Then, the ADE and FDE in meters are computed as:

\begin{equation}
\mathrm{ADE}_\text{meters} = \frac{1}{T} \sum_{t=1}^{T} d_t, \quad
\mathrm{FDE}_\text{meters} = d_T
\end{equation}

These metrics provide a rigorous and interpretable way to evaluate how closely the model's predicted trajectory aligns with the actual observed iceberg movement, both in spherical coordinates and great circle distance calculated using Equation \ref{eq: geodesic_distance}. To obtain the compact value, we further converted this meters further into kilometers that are highlighted in the result section.

\subsection{Result Analysis}
\begin{figure*}[t]
  \centering
  \includegraphics[width=1\textwidth]{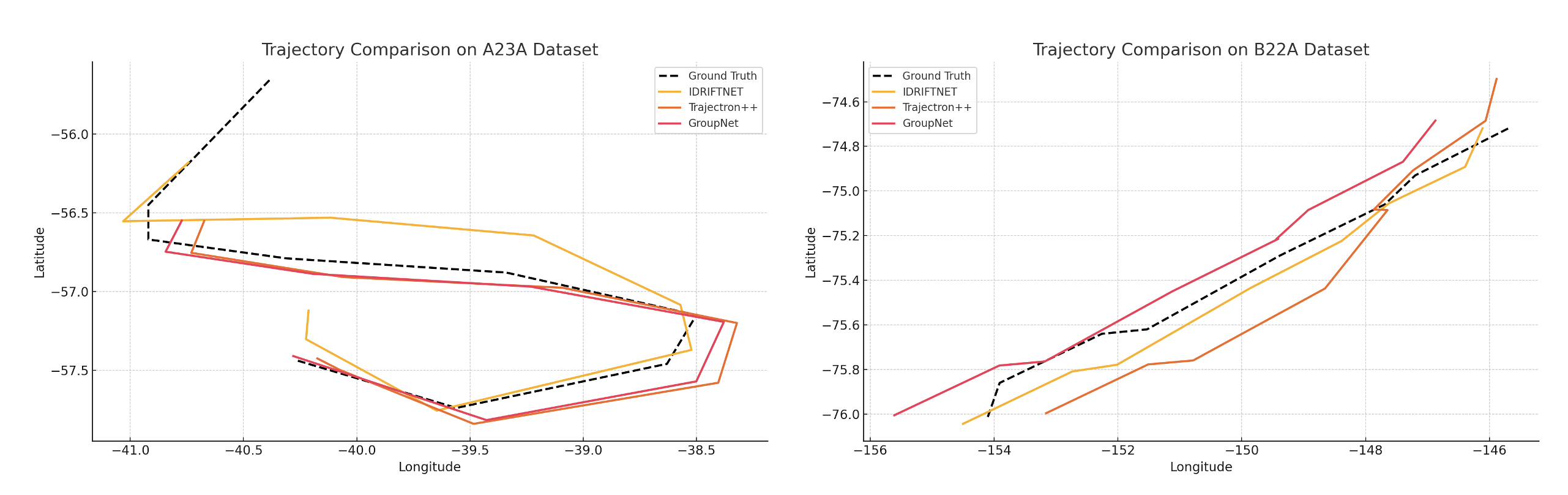}
  \caption{Comparison of the top-3 trajectory prediction models on A23A (left) and B22A (right). Ground truth iceberg trajectories are shown with black dashed lines. IDRIFTNET  consistently tracks the true drift path with high spatial accuracy. Trajectron++ and GroupNet  exhibit larger deviations, particularly in nonlinear and long-range drift segments.}
  \label{fig:top3_trajectory_comparison}
\end{figure*}

Table 1 presents a comprehensive evaluation of ten physics-integrated trajectory forecasting models tested on the A23A iceberg dataset. To ensure a fair comparison, all models have been adapted to include a residually connected physics-based drift module following the formulation of Wagner et al. (2017). This harmonizes the physical inductive bias across all baselines, allowing the performance differences to primarily reflect the capabilities of their respective learning architectures. Each model is evaluated using two widely adopted metrics: Average Displacement Error (ADE) and Final Displacement Error (FDE), computed in both Euclidean coordinates (degrees) and geodesic distances (kilometers), with the latter calculated using the WGS84 ellipsoid. While Euclidean metrics provide angular displacement estimates, geodesic distances offer a physically meaningful interpretation of spatial accuracy on the Earth's surface. Among these, FDE (Geodesic) serves as a particularly critical metric for long-range prediction, as it quantifies the final spatial deviation of the predicted iceberg position from the observed trajectory.

The proposed model, IDRIFTNET, outperforms all competitors, achieving the lowest geodesic FDE of 63.32 km and ADE of 49.06 km. This highlights its superior ability to model both intermediate and final positions accurately. IDRIFTNET incorporates a hybrid architecture combining Gabor spectral encoding and physics-informed residual drift modeling. The Gabor module captures multi-scale temporal features with improved frequency localization, while the physics-based residual term ensures consistency with real-world environmental dynamics such as wind and ocean current forces. This combination enables robust generalization across sparse and noisy temporal inputs.

TRAJECTRON++~\cite{salzmann2020trajectron++} ranks second, yielding a geodesic FDE of 102.10 km and ADE of 56.26 km. Though originally designed for multimodal pedestrian behavior, its modular encoder-decoder structure—augmented with a physical residual—retains some efficacy in drift modeling. However, the lack of strong spatial inductive bias results in higher endpoint error compared to IDRIFTNET. GROUPNET~\cite{xu2022groupnet} also performs competitively, with a geodesic FDE of 103.18 km and ADE of 53.17 km, outperforming PECNet, NPSN, and diffusion-based models. Its multi-scale hypergraph reasoning captures group-wise motion patterns more effectively than traditional relational models.

AGENTFORMER~\cite{yuan2021agentformer} records an FDE of 99.96 km and ADE of 51.83 km, maintaining strong performance but originally designed for social interactions rather than force-driven dynamics. PECNet~\cite{mangalam2020not}, despite its endpoint-conditioned sampling and added physical priors, achieves a much higher FDE of 294.41 km. Its latent endpoint estimation introduces variance, particularly under sparse observational data, making it less suitable for environmental modeling. Similarly, NPSN~\cite{bae2022non} performs moderately with 369.28 km FDE but suffers from stochastic sampling inconsistencies even with QMC sampling. Graph-based methods such as NMMP~\cite{hu2020collaborative} and STAR~\cite{yu2020spatio} record geodesic FDEs of 16171.56 km and 3058.68 km, respectively. Their message-passing and spatiotemporal attention mechanisms fail to model large-scale passive motion governed by physical laws. Diffusion-based models like LEAPFROG~\cite{mao2023leapfrog} and MID~\cite{gu2022stochastic} produce the worst results with FDEs of 2817.31 km and 2571.54 km, reflecting a lack of trajectory anchoring despite improved diversity in generated paths.

Overall, IDRIFTNET achieves the best geodesic performance on both ADE and FDE. In terms of FDE (GEO), it outperforms TRAJECTRON++ (102.10 km) by 38.0\%, GROUPNET (103.18 km) by 38.7\%, and PECNet (294.41 km) by 78.5\%. This demonstrates IDRIFTNET’s effectiveness in fusing temporal encoding and physical realism for accurate, long-horizon iceberg trajectory prediction.

\begin{table}[H]
\centering
\caption{A23A Trajectory Error Comparison. ADE: Average Displacement Error, FDE: Final Displacement Error. EUC: Euclidean (in degrees). GEO: Geodesic in kilometers computed using the WGS84 ellipsoid.}
\resizebox{\linewidth}{!}{
\begin{tabular}{@{}lcccc@{}}
\toprule
\textbf{Method} & \textbf{ADE (EUC)} & \textbf{FDE (EUC)} & \textbf{ADE (km)} & \textbf{FDE (km)} \\
\midrule
AGENTFORMER \cite{yuan2021agentformer}        & 0.71     & 1.00     & 51.83   & 99.96 \\
LEAPFROG \cite{mao2023leapfrog}               & 27.89    & 28.79    & 2678.99 & 2817.31 \\
MID \cite{gu2022stochastic}                   & 28.26    & 27.31    & 2647.25 & 2571.54 \\
NMMP \cite{hu2020collaborative}               & 148.60   & 201.76   & 12787.26& 16171.56 \\
NPSN \cite{bae2022non}                        & 5.06     & 5.53     & 352.39  & 369.28 \\
PECNET \cite{mangalam2020not}                 & 3.62     & 2.89     & 268.92  & 294.41 \\
STAR \cite{yu2020spatio}                      & 32.72    & 55.45    & 2024.24 & 3058.68 \\
TRAJECTRON++ \cite{salzmann2020trajectron++}  & 0.72     & 0.94     & 56.26   & 102.10 \\
GROUPNET \cite{xu2022groupnet}                & 0.67     & 0.98     & 53.17   & 103.18 \\
\textbf{IDRIFTNET (Ours)}                     & \textbf{0.65} & \textbf{0.64} & \textbf{49.06} & \textbf{63.32} \\
\bottomrule
\end{tabular}
}
\end{table}

\vspace{0.2cm}
Table 2 summarizes the trajectory prediction results across different state-of-the-art models on the B22A iceberg dataset using both Euclidean and geodesic metrics. Of particular importance is the Final Displacement Error (FDE) in geodesic space, as it reflects the model's ability to predict the final position of the iceberg with high spatial fidelity—a critical requirement for real-world polar applications.

Among all evaluated methods, the proposed IDRIFTNET achieves the best performance with a geodesic FDE of 10.87 km and an ADE of 22.86 km, outperforming all other state-of-the-art baselines by a significant margin. This superior accuracy stems from its architecture that explicitly integrates physics-based drift modeling with a Gabor-based spectral residual decoder, enabling it to effectively model complex, long-range drift dynamics under nonlinear environmental influences.

In contrast, the second-best performer, TRAJECTRON++~\cite{salzmann2020trajectron++}, records a geodesic FDE of 25.82 km and ADE of 40.89 km. While Trajectron++ is designed for multimodal predictions in dynamic environments, it lacks strong inductive bias toward physical motion, making it less optimal in scenarios like iceberg drift that are driven by deterministic environmental forces. GroupNet~\cite{xu2022groupnet}, despite leveraging multiscale hypergraph reasoning to model pairwise and group-wise interactions, ranks third with an FDE of 40.52 km. Its design prioritizes relational dynamics over environmental modeling, limiting its generalization to real-world drift phenomena.

AgentFormer~\cite{yuan2021agentformer}, which models temporal and social dependencies via agent-aware attention mechanisms, records an FDE of 429.98 km. While it excels in human-agent interaction tasks, its generalization degrades significantly in physically governed domains such as polar drift. Generative and stochastic models like LEAPFROG~\cite{mao2023leapfrog} and MID~\cite{gu2022stochastic} yield geodesic FDEs of 234.51 km and 71.10 km, respectively. Though their denoising diffusion strategies support multimodal predictions, they often result in high variance and low stability under geophysical constraints.

Graph-based models like NMMP~\cite{hu2020collaborative} and NPSN~\cite{bae2022non} struggle to incorporate physical priors despite their relational modeling capacities. Their geodesic FDEs—18442.75 km and 237.58 km, respectively—highlight their poor suitability for iceberg prediction, especially in sparse or low-resolution datasets. Transformer-based models such as STAR~\cite{yu2020spatio}, with an FDE of 2941.76 km, also overfit in low-data regimes and fail to encode spatial inductive biases critical for oceanic dynamics. Likewise, endpoint-conditioned models like PECNet~\cite{mangalam2020not} reach an FDE of 481.73 km, further reinforcing the need for physically grounded architectures.

The top three models by geodesic FDE are: IDRIFTNET (10.87 km), TRAJECTRON++ (25.82 km), and GroupNet (40.52 km). Compared to GroupNet, IDRIFTNET achieves a 73.2\% reduction in FDE, and compared to Trajectron++, it reduces FDE by 57.9\%. These results confirm that domain-specific hybrid modeling anchored in physics and augmented with spectral learning is essential for accurate and reliable prediction of the iceberg trajectory.

The results from Tables 1 and 2 demonstrate that IDRIFTNET consistently outperforms all baseline models across both iceberg datasets, A23A and B22A. In each case, it achieves the lowest displacement errors, both in terms of average and final distances, using geodesic and Euclidean measures. These outcomes highlight IDRIFTNET’s ability to maintain accuracy in predicting long-term iceberg trajectories, even under limited and noisy observational conditions.

What distinguishes IDRIFTNET is its hybrid design that brings together principles from physics and data-driven learning. At the core of the model is a physics-informed module that estimates drift velocity using external forces such as wind and ocean current.Complementing this is a spectral encoder-decoder architecture based on Gabor filters. Unlike standard temporal models, this spectral module transforms the input sequence into the frequency domain, allowing it to capture smooth trends, periodic shifts, and long-range temporal dependencies more effectively. Together, these two components work in synergy: the physics module ensures the model respects environmental dynamics, while the spectral block improves the model’s temporal generalization.
\begin{table}[H]
\centering
\caption{B22A Trajectory Error Comparison. ADE: Average Displacement Error, FDE: Final Displacement Error. EUC: Euclidean (in degrees). GEO: Geodesic in kilometers computed using the WGS84 ellipsoid.}
\resizebox{\linewidth}{!}{
\begin{tabular}{@{}lcccc@{}}
\toprule
\textbf{Method} & \textbf{ADE (EUC)} & \textbf{FDE (EUC)} & \textbf{ADE (km)} & \textbf{FDE (km)} \\
\midrule
AGENTFORMER \cite{yuan2021agentformer}        & 12.11     & 12.83   & 396.16   & 429.98 \\
LEAPFROG \cite{mao2023leapfrog}               & 8.82     & 7.70    & 269.93  & 234.51 \\
MID \cite{gu2022stochastic}                   & 23.34    & 27.01   & 647.28   & 71.10 \\
NMMP \cite{hu2020collaborative}               & 76.07     & 168.02  & 7604.91  & 18442.75 \\
NPSN \cite{bae2022non}                        & 3.41     & 5.47     & 178.08   & 237.58 \\
PECNET \cite{mangalam2020not}                 & 11.63    & 15.74   & 359.42   & 481.73 \\
STAR \cite{yu2020spatio}                      & 54.62    & 106.91  & 1490.42  & 2941.76 \\
TRAJECTRON++ \cite{salzmann2020trajectron++}    & 1.28     & 0.95    & 40.89    & 25.82 \\
GROUPNET \cite{xu2022groupnet}                & 0.82     & 1.50    & 26.81    & 40.52 \\
\textbf{IDRIFTNET (Ours)}                                & \textbf{0.72} & \textbf{0.39} & \textbf{22.86} & \textbf{10.87} \\
\bottomrule
\end{tabular}
}
\end{table}

To understand the value of hybrid learning, we compare the performance of IDRIFTNET with a physics-only baseline that uses Wagner et al. 2017's analytical drift formulation. This baseline estimates iceberg motion based solely on environmental inputs such as wind, ocean currents, and area, without any learning component or adaptive correction.

Table 3 summarizes the results on both the A23A and B22A datasets, comparing the purely physics-based prediction proposed in Wagner et al. 2017 \cite{wagner2017analytical} with our proposed IDRIFTNET model. On A23A, the physics-only model achieves a geodesic ADE of 147.10 km and an FDE of 117.54 km. In contrast, IDRIFTNET reduces these errors to 49.06 km and 63.32 km respectively, demonstrating a substantial improvement in both intermediate and final trajectory accuracy. In Euclidean terms, IDRIFTNET lowers the ADE from 1.69 to 0.65 and the FDE from 1.81 to 0.64.The performance difference is also significant on the B22A dataset. The physics-only model produces a geodesic ADE of 127.23 km and FDE of 39.50 km, while IDRIFTNET achieves 22.87 km and 10.87 km respectively. This reflects a strong gain in both spatial tracking and final displacement accuracy. Similarly, the Euclidean errors are reduced from 4.01 to 0.72 for ADE, and from 0.67 to 0.39 for FDE. These results highlight that while the physics-based model captures general drift behavior well, it is limited by its fixed structure and cannot adapt to complex variations in the observed data. IDRIFTNET addresses this by combining the physics-based output with a learnable residual component, enabling it to capture deviations due to unmodeled dynamics, localized anomalies, or changes in environmental response.

\begin{table}[H]
\centering
\caption{Comparison of the physics-only Wagner et al. 2017 model (TW) and IDRIFTNET on A23A and B22A datasets.}
\label{tab:physics_vs_hybrid}
\resizebox{\linewidth}{!}{
\begin{tabular}{@{}clcccc@{}}
\toprule
\textbf{Dataset} & \textbf{Method} & \textbf{ADE (EUC)} & \textbf{FDE (EUC)} & \textbf{ADE (km)} & \textbf{FDE (km)} \\
\midrule
A23A & Physics-Only (TW)        & 1.69  & 1.81  & 147.10   & 117.54   \\
A23A & IDRIFTNET (Ours)         & \textbf{0.65} & \textbf{0.64} & \textbf{49.06} & \textbf{63.32} \\
B22A & Physics-Only (TW)        & 4.01  & 0.67   & 127.23   & 39.50   \\
B22A & IDRIFTNET (Ours)         & \textbf{0.72} & \textbf{0.39} & \textbf{22.87} & \textbf{10.87}  \\
\bottomrule
\end{tabular}
}
\end{table}

\begin{figure*}[t]
  \centering
  \includegraphics[width=1.\textwidth]{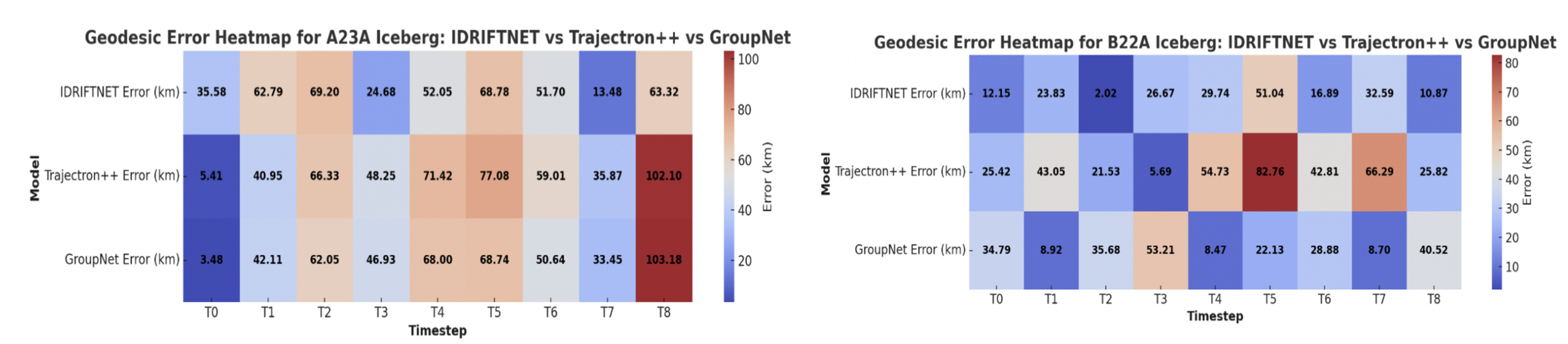}
  \caption{Geodesic error heatmaps comparing the top-3 trajectory prediction models—IDRIFTNET, Trajectron++, and GroupNet—for A23A (left) and B22A (right) icebergs.}
  
\end{figure*}

To evaluate the benefit of including physics-based  residual modeling in our approach, we conducted an ablation study using two large iceberg datasets: A23A and B22A. Table~\ref{tab:ablation_idriftnet_trajectron} presents the results for IDRIFTNET and Trajectron++ with and without residual connections to the physics module. We report both Euclidean errors (which measure coordinate-level differences) and geodesic errors (which reflect actual distances on Earth’s surface in kilometers). Removing the residual physics connection significantly increased error in all cases. For example, in the A23A dataset, IDRIFTNET's geodesic ADE increased from 49.06 km to 341.25 km, and FDE from 63.32 km to 328.65 km. In the B22A dataset, ADE rose from 22.87 km to 679.70 km, and FDE from 10.87 km to 804.34 km. A similar pattern was observed with Trajectron++, where the absence of residual physics also led to much higher prediction errors.These results show that connecting physics-based modeling with deep learning improves both short-term and final trajectory predictions. In total, the integration of physics via residual learning, as implemented in IDRIFTNET, not only stabilizes training but significantly improves forecasting accuracy. It is particularly effective in regimes like iceberg drift, where geophysical forces are dominant and annotated data is limited.

\begin{table}[H]
\centering
\caption{Ablation study performance comparison of IDRIFTNET and Trajectron++ on A23A and B22A glaciers with and without residual physics connection. GEO values are shown in kilometers.}
\label{tab:ablation_idriftnet_trajectron}
\resizebox{\linewidth}{!}{
\begin{tabular}{@{}lcccc@{}}
\toprule
\textbf{Method Name} & \textbf{ADE (EUC)} & \textbf{FDE (EUC)} & \textbf{ADE (GEO)} & \textbf{FDE (GEO)} \\
\midrule
IDRIFTNET: A23A without Res. Connected Physics       & 5.19  & 5.06  & 341.25  & 328.65  \\
\textbf{IDRIFTNET: A23A with Res. Connected Physics(Ours)} & \textbf{0.65} & \textbf{0.64} & \textbf{49.06}   & \textbf{63.32}   \\
IDRIFTNET: B22A without Res. Connected Physics       & 22.88 & 27.39 & 679.70  & 804.34  \\
\textbf{IDRIFTNET: B22A with Res. Connected Physics(Ours)} & \textbf{0.72} & \textbf{0.39} & \textbf{22.87}   & \textbf{10.87}   \\
\midrule
Trajectron++: A23A without Res. Connected Physics     & 5.65  & 8.00  & 403.02  & 658.85  \\
Trajectron++: A23A with Res. Connected Physics        & 0.72  & 0.94  & 56.26  & 102.10 \\
Trajectron++: B22A without Res. Connected Physics     & 23.11 & 28.40 & 692.65  & 836.46 \\
Trajectron++: B22A with Res. Connected Physics        & 1.28  & 0.95  & 40.89  & 25.82  \\
\bottomrule
\end{tabular}
}
\end{table}

\begin{table}[H]
\centering
\caption{Ablation study on IDRIFTNET for the A23A and B22A glaciers, evaluating the impact of removing Rotate Block and Gabor Filtering. Geodesic values (GEO) are shown in kilometers.}
\label{tab:ablation_rotate_gabor}
\resizebox{\linewidth}{!}{
\begin{tabular}{@{}lcccc@{}}
\toprule
\textbf{Method Name} & \textbf{ADE (EUC)} & \textbf{FDE (EUC)} & \textbf{ADE (GEO)} & \textbf{FDE (GEO)} \\
\midrule
IDRIFTNET: A23A without Rotate Block        & 0.71  & 0.82  & 53.32  & 88.43  \\
IDRIFTNET: A23A without Gabor Filtering     & 0.91  & 0.97  & 70.13  & 108.45 \\
\textbf{IDRIFTNET (Ours): A23A}             & \textbf{0.65} & \textbf{0.64} & \textbf{49.06} & \textbf{63.32} \\
\midrule
IDRIFTNET: B22A without Rotate Block        & 0.74  & 0.56  & 30.92  & 25.57  \\
IDRIFTNET: B22A without Gabor Filtering     & 1.21  & 2.05  & 46.13  & 62.36  \\
\textbf{IDRIFTNET (Ours): B22A}             & \textbf{0.72} & \textbf{0.39} & \textbf{22.87} & \textbf{10.87} \\
\bottomrule
\end{tabular}
}
\end{table}

To further analyze the importance of the architectural components of IDRIFTNET, we conducted an ablation study focusing on two critical elements: the Rotate Block and Gabor spectral filtering. As shown in Table 5, removing the Rotate Block led to a moderate decline in performance, with geodesic ADE increasing from baseline levels to 53.32 km and 30.92 km on A23A and B22A respectively. However, the removal of Gabor filtering resulted in a more significant degradation, especially for the A23A glacier, where the geodesic FDE rose to 108.45 km. This trend was also evident in the B22A case, where removing Gabor layers led to ADE and FDE increases to 46.13 km and 62.36 km respectively as compared to our proposed approach. These results confirm that both the Rotate Block and Gabor filtering contribute meaningfully to IDRIFTNET’s ability to capture spatiotemporal dependencies, with Gabor-based spectral encoding playing a particularly crucial role in modeling trajectory dynamics across different glaciological contexts.
.

\subsection{Visualization}
To qualitatively and quantitatively assess the predictive capabilities of the evaluated models—IDRIFTNET, GroupNet, and Trajectron++  as shown in Figures 3 and 4, we presented two complementary visualization strategies. These include spatial trajectory overlays (Figure 3) and geodesic error heatmaps (Figure 4), both analyzed for two iceberg cases: A23A and B22A.

Figure 3 illustrates the spatiotemporal alignment of predicted iceberg drift trajectories overlaid against the ground truth paths for both A23A and B22A. The trajectories are represented as polylines in longitude-latitude space. The dashed black line corresponds to the observed trajectory (ground truth), while solid lines denote predictions from the three models—yellow for IDRIFTNET, orange for Trajectron++, and red for GroupNet. For the A23A iceberg, the observed path shows a complex pattern with a critical directional change in the mid-sequence. IDRIFTNET closely follows this curvature, preserving spatial consistency across the segment. In contrast, GroupNet exhibits misalignment in earlier segments and diverges significantly near the turn, while Trajectron++ fails to model the nonlinear deviation and shows trajectory discontinuities, suggesting weaker generalization to abrupt shifts. In the B22A iceberg scenario, which demonstrates a smoother, more linear drift pattern at this stage, IDRIFTNET again tracks the ground truth closely throughout the sequence. GroupNet’s predictions show higher variance across mid-trajectory points, while Trajectron++ increasingly drifts after the midpoint. These spatial patterns visually confirm IDRIFTNET’s robustness across complex and regular regimes. The corresponding quantitative deviations are clearly shown in Figure 4, offering a more fine-grained temporal breakdown of geodesic errors.

Figure 4 presents a heatmap of geodesic errors (in kilometers) for each model at each timestep (T0–T8), providing a fine-grained, quantitative view of trajectory accuracy. For both A23A and B22A, IDRIFTNET consistently achieves the lowest errors across nearly all timesteps. In the A23A case, geodesic errors for IDRIFTNET remain below 70 km, with notable dips at T0 (35.58 km) and T7 (13.48 km). In contrast, Trajectron++ and GroupNet record significantly higher errors during turning regions, peaking at over 100 km by T8. This supports the hypothesis that IDRIFTNET’s hybrid structure, combining spectral learning and physics-based residuals, enables effective modeling of abrupt transitions. In the B22A case, where geodesic drift is more gradual, IDRIFTNET maintains low error magnitudes under 20 km across most intervals, reaching as low as 2.02 km at T2. Trajectron++ shows higher variability, with errors spiking to 82.76 km at T5, while GroupNet also demonstrates inconsistency, particularly at T3 (53.21 km) and T8 (40.52 km). These results quantitatively reaffirm the spatial patterns observed in the trajectory plots and demonstrate IDRIFTNET’s stability and adaptability across different iceberg movement regimes.

Collectively, these visualizations reinforce IDRIFTNET’s superior performance, both in spatial fidelity and quantitative accuracy. The integration of domain-informed physics with learned spectral-temporal encoders allows it to anticipate both smooth drifts and nonlinear deviations. This makes IDRIFTNET especially suitable for real-world forecasting tasks involving polar dynamics, where uncertainty, curvature, and temporal heterogeneity must be accurately captured.

\subsection{Model Specifications}
Table 6 highlights that our multistage development of IDRIFTNET not only led to substantial accuracy gains but also delivered computational efficiency compared to state-of-the-art models. Specifically, IDRIFTNET consistently achieves the lowest Final Displacement Error (FDE) across both the A23A and B22A datasets, demonstrating superior predictive accuracy compared to all other evaluated models.

Importantly, these predictive improvements are achieved with only 0.29 million trainable parameters and 1.14~MB of memory usage—significantly lower than large models such as AgentFormer (7.16M, 27.38~MB) and PECNet (1.40M, 5.40~MB). While IDRIFTNET has slightly more parameters than minimal baselines such as NMMP and STAR, it offers far superior accuracy, demonstrating strong performance-per-parameter efficiency.

This favorable balance between accuracy and model compactness underscores the architectural efficiency of IDRIFTNET, making it not only more accurate but also lightweight and well-suited for deployment in resource-constrained or real-time forecasting environments.

\begin{table}[H]
\centering
\caption{Model specifications in terms of trainable parameters (in millions) and memory size (in MB).}
\label{tab:model_specifications}
\resizebox{\linewidth}{!}{
\begin{tabular}{@{}lcc@{}}
\toprule
\textbf{Method} & \textbf{Trainable Parameters (Mill.)} & \textbf{Memory Size (MB)} \\
\midrule
AGENTFORMER \cite{yuan2021agentformer}     & 7.16  & 27.38 \\
LEAPFROG \cite{mao2023leapfrog}            & 0.12  & 0.50  \\
MID \cite{gu2022stochastic}                & 0.38  & 1.50  \\
NMMP \cite{hu2020collaborative}            & 0.02  & 0.13  \\
NPSN \cite{bae2022non}                     & 0.18  & 0.77  \\
PECNET \cite{mangalam2020not}              & 1.40  & 5.40  \\
STAR \cite{yu2020spatio}                   & 0.045 & 0.26  \\
TRAJECTRON++ \cite{salzmann2020trajectron++} & 0.15  & 0.61  \\
GROUPNET \cite{xu2022groupnet}             & 0.20  & 0.83  \\
\textbf{IDRIFTNET (Ours)}                  & \textbf{0.29}  & \textbf{1.14}  \\
\bottomrule
\end{tabular}
}
\end{table}

\section{Conclusion and Future Work}
This paper presents IDRIFTNET, a hybrid deep learning model that combines physics-based drift modeling with spectral-temporal learning to forecast iceberg trajectories. By integrating an analytical formulation of iceberg dynamics with a rotated  Gabor Spectral Network, the model captures both physically driven motion and complex temporal patterns. The spectral module enables frequency-domain learning, effectively modeling long-range dependencies and sudden drift shifts, while the residual physics module grounds predictions in known environmental forces. Across both A23A and B22A datasets, IDRIFTNET outperforms baseline models, achieving the lowest displacement errors in both average and final metrics. Visual and quantitative results confirm that the model maintains accuracy across diverse drift regimes, including nonlinear turns regions. Ablation studies further highlight the importance of both physics-informed initialization, rotate block importance, and gabor spectral correction. Despite its strong performance, further evaluation is needed. Future work will focus on testing IDRIFTNET on a wider set of icebergs with varying geometries and environmental conditions to assess robustness and transferability. Expanding to new datasets will help validate the model’s generalization and guide future improvements, with the ultimate goal of enabling scalable, reliable iceberg monitoring in polar regions.
\section{Acknowledgments}
This work is funded by the NSF award 2118285.

\bibliographystyle{unsrt}
\bibliography{references}

\end{document}